%% file: main.tex
\pdfoutput=1

\documentclass[11pt]{article}

\usepackage{ACL2023}

\usepackage{times}
\usepackage{latexsym}
\usepackage{enumitem}

\usepackage{graphicx}
\usepackage{adjustbox}
\usepackage{array}
\setlist{nosep}
\usepackage{microtype}
\usepackage{graphicx}
\usepackage{booktabs} 
\usepackage{amsmath}
\usepackage{amsfonts}
\usepackage{subcaption}
\usepackage{multirow}
\usepackage{xcolor}

\usepackage[T1]{fontenc}

\usepackage[utf8]{inputenc}

\usepackage{microtype}

\usepackage{inconsolata}
\newcommand{\email}[1]{\textcolor{blue}{\small\texttt{\href{mailto:#1}{#1}}}}

\title{What if Red Can Talk? Dynamic Dialogue Generation Using Large Language Models}

\author{Navapat Nananukul \\
  University of Southern California \\
  \email{nananuku@usc.edu} \\\And
  Wichayaporn Wongkamjan \\
  University of Maryland \\
  \email{wwongkam@umd.edu} \\}

\begin{document}
\maketitle
\input{sections/00-abstract}
\input{sections/20-games}

\input{sections/30-method}

\input{sections/40-results}
\input{sections/50-related}

\input{sections/60-conclusion}

\bibliography{anthology,custom}
\bibliographystyle{acl_natbib}

\end{document}

%% file: sections/00-abstract.tex
\begin{abstract}

Role-playing games (RPGs) provide players with a rich, interactive world to explore. Dialogue serves as the primary means of communication between developers and players, manifesting in various forms such as guides, NPC interactions, and storytelling. While most games rely on written scripts to define the main story and character personalities, player immersion can be significantly enhanced through casual interactions between characters. With the advent of large language models (LLMs), we introduce a dialogue filler framework that utilizes LLMs enhanced by knowledge graphs to generate dynamic and contextually appropriate character interactions. We test this framework within the environments of Final Fantasy VII Remake and Pokémon, providing qualitative and quantitative evidence that demonstrates GPT-4’s capability to act with defined personalities and generate dialogue. However, some flaws remain, such as GPT-4 being overly positive or more subtle personalities, such as maturity, tend to be of lower quality compared to more overt traits like timidity. This study aims to assist developers in crafting more nuanced filler dialogues, thereby enriching player immersion and enhancing the overall RPG experience.

\end{abstract}

%% file: sections/20-games.tex
\section{Human Assistance --- RPG-dialogue Large Language Model}
\label{sec:games}

Role-playing games have been identified as a significant challenge for the next stage of artificial intelligence research. These games present a complex, open-ended creative arena where players interact through dialogue to shape the narrative and game outcomes. Unlike traditional games such as Chess and Go, which are zero-sum and deterministic.

\begin{figure}[t]
    \centering
    \includegraphics[width=0.49\textwidth]{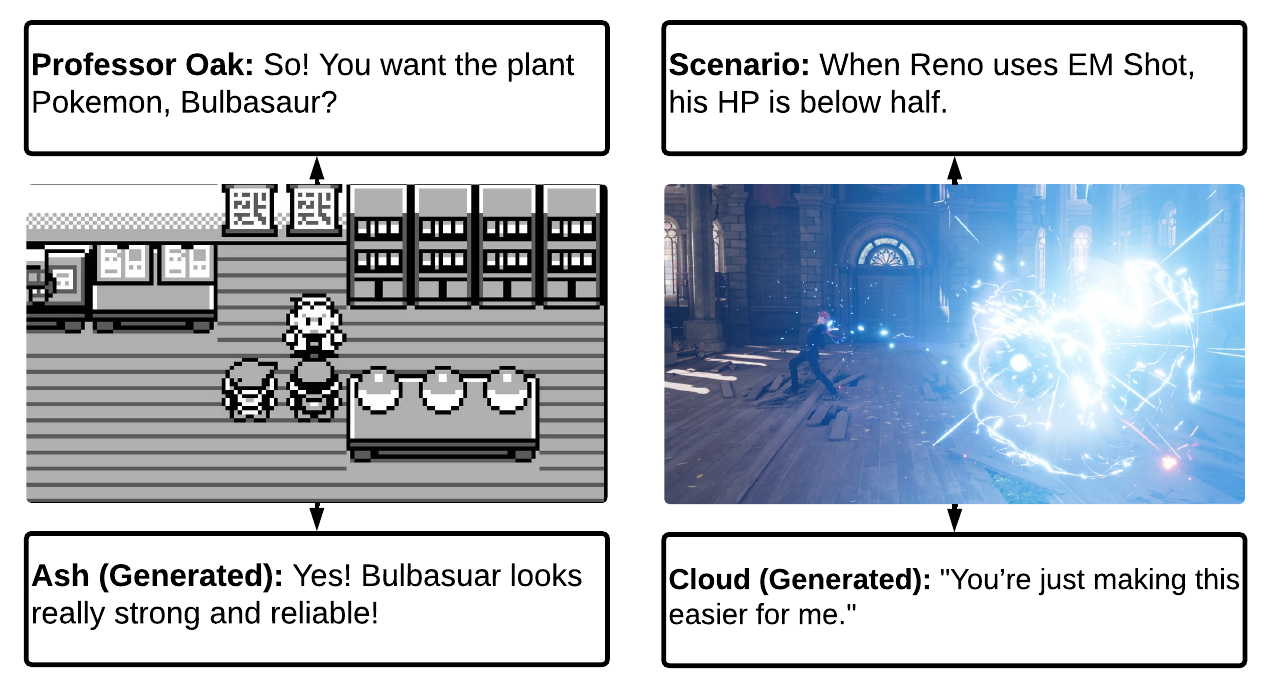}
    \caption{An illustration example of LLMs-generated dialogue in Pokémon (Left) and Final Fantasy VII Remake (FFVIIR). Left panel: An unseen dialogue interaction in Pokémon, where Red selected his first Pokémon Bulbusuar. Right panel: A scenario from Final Fantasy VII where Cloud responds during combat. }
    \label{fig:motivating_example}
\end{figure}

One of the critical challenges presented by role-playing games is the necessity for nuanced, context-aware dialogue. Players create characters that interact within a fictional setting, making decisions and taking actions that influence the game's progression. One example is Dungeon and Dragons (DND) where the game's facilitator, often known as the Dungeon Master (DM) or Game Master (GM), describes scenarios, locations, and the roles of non-player characters (NPCs) and monsters. The DM's role highlights the importance of narrative and improvisation, further complicating the AI's task.

Another challenges in RPG dialogue is the occurrence of unnatural silences from main characters or NPCs. In real-world interactions, conversations flow continuously, with participants actively contributing to the dialogue. However, in RPGs, there can be instances where characters remain silent, breaking the immersion and disrupting the natural flow of the narrative. 

These challenges become even more critical in popular games where main characters have strong personalities. For example, in \texttt{Final Fantasy VII Remake (FFVIIR)}, one of the most popular story-based RPGs, the main character \textit{``Cloud Strife''} is known for his brooding and complex personality, characterized by his introspective nature and struggle with his identity. Another example is \texttt{Pokémon}, where the main character, \textit{``Ash Ketchum''} (or Red in the original games), exhibits a determined and optimistic personality, always eager to engage with others and face new challenges. The inconsistency of dialogue, especially moments of unnatural silence, can disrupt the portrayal of these rich, well-defined characters, breaking player immersion and diminishing the storytelling experience. Addressing this challenge requires advanced AI capable of maintaining continuous, contextually appropriate dialogue that aligns with each character's established traits, ensuring a more engaging and authentic player experience.

With the advent of AI and large language models (LLMs), role-playing games (RPGs) have become valuable testbeds due to their ability to simulate complex real-world dynamics, such as language generation, knowledge acquisition, and strategic planning tailored to each character and game scenario. This evolution presents developers with the opportunity to leverage these advances as tools for generating nuanced filler dialogues (Figure \ref{fig:motivating_example}). Our contributions in this paper include:

\begin{itemize}
    \item \textbf{Integration of Knowledge Graphs and LLMs for Dialogue Generation:} We built game-specific knowledge graph and combined knowledge graphs with large language models (LLMs) to produce dialogues that are not only contextually rich but also character-specific, thus providing character-specific information to dialogue generation step.
    
    \item \textbf{Evaluate the Capability of Current LLMs Technology:} We assessed how well current version of GPT-4, when integrated with knowledge-based systems, can generate high-quality game dialogues that contribute to a more engaging and interactive narrative.
    
\end{itemize}

%% file: sections/30-method.tex
\begin{figure*}[ht]
    \centering
    \includegraphics[width=\textwidth]{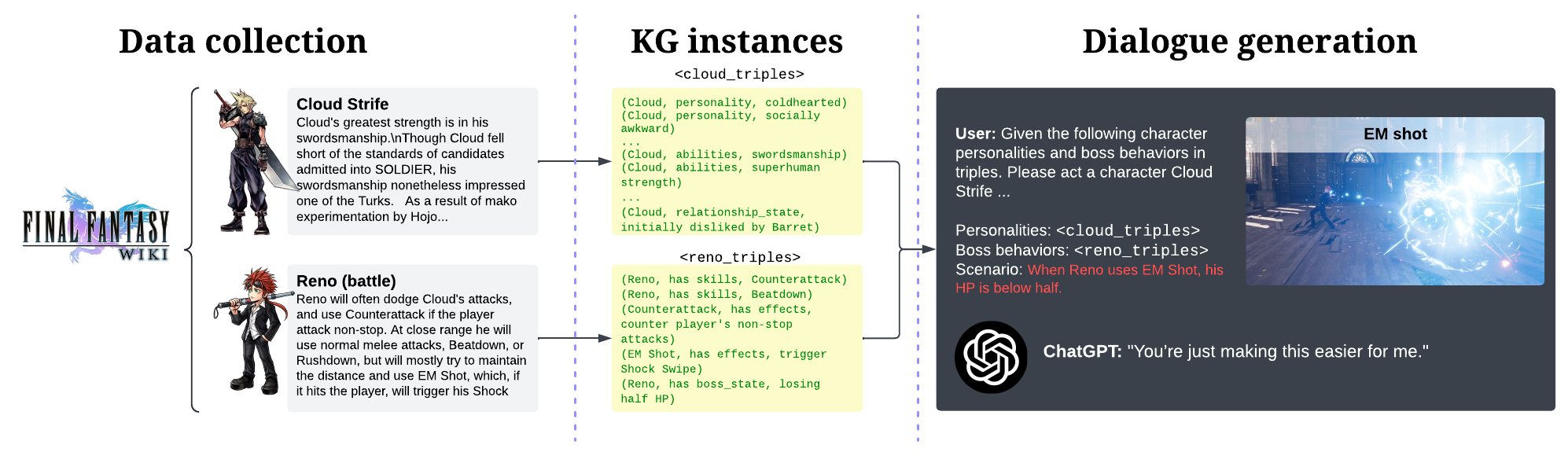}
    \caption{An illustrative examples of our method using Final Fantasy VII Remake (FFVIIR) as a games example. The process begins with data collection, where character information is extracted from FFVII wiki pages. Subsequently, knowledge graph (KG) instances are generated using GPT-4 alongside predefined ontology concepts and relations, creating a game-specific KG. Finally, GPT-4 is prompted with actual game dialogue integrated with knowledge from the KG, enabling contextually enriched dialogue generation.}
    \label{fig:steps}
\end{figure*}

\section{Dialogue Generation and Completion}
\label{sec:dialogue}
As discussed in Section~\ref{sec:games}, in-game dialogue plays a crucial role in story-driven RPGs. 
While some dialogues can be improved, others can be generated anew.
We believe that computational resources can complement human efforts in creating high-quality dialogue. 
By leveraging advanced algorithms and AI, we aim to produce dialogues that match the quality of those written by humans, thereby optimizing our resources and enhancing the overall player experience.

To assist humans in dialogue-related tasks, we focus on addressing specific challenges encountered in two different games: \texttt{FFVIIR} and \texttt{Pokémon}.
One notable challenge in \texttt{FFVIIR} is the repetitive in-battle dialogue, which lacks variation and fails to adapt to the boss's changing states. 
For example, the main characters, including \textit{``Cloud''}, often remain silent when a boss is low on health and nearly defeated.
We aim to improve the player experience by generating dynamic dialogue for the main characters that respond to different battle states. 
This would allow for more spontaneous and interactive interpersonal communication during battles.
In the original \texttt{Pokémon} series, the main character, Red, does not say a word during gameplay. 
This piques our curiosity and inspires us to create dialogues for \textit{``Red''} to respond in different situations.
Since Red is not explicitly defined by the Pokemon games, he serves as a blank canvas that we can enrich with various characteristics.
For instance, if Red is portrayed as hot-headed, he might yell at his Pokémon during intense moments. Conversely, if he is depicted as kind-hearted, he would thank his Pokémon when a battle concludes.

With the rise of LLMs, we consider GPT-4 \citep{openai2024gpt4} to generate dialogues where character profiles, NPC backgrounds, and scenarios are given.
While we could simply input all relevant details about characters, NPCs, bosses, or Pokémon into a prompt and ask GPT-4 to generate dialogue, studies have shown that incorporating knowledge-based data, such as knowledge graphs, can enhance text generation quality \cite{pan2024unifying}.
We hereby propose a framework to generate in-game dialogue based on given knowledge.
First, we scrape data from online sources, Final Fantasy Wiki\footnote{Link to website: \href{https://finalfantasy.fandom.com/wiki/Final_Fantasy_Wiki}{Final Fantasy Wiki}} and Pokémon Wiki\footnote{Link to website: \href{https://pokemon.fandom.com/wiki/Pok\%C3\%A9mon_Wiki}{Pokémon Wiki}} (left, Figure~\ref{fig:steps}).
Next, we create triples for knowledge graph instances (middle, Figure~\ref{fig:steps}). 
This step can be done manually by experts or generated by GPT-4\cite{biswas2021contextual} .
Since we have a short time frame, we prompt GPT-4 with a defined ontology specifically for each game; then, we hand-pick triples that are reasonable to represent entities (More details in Section~\ref{sec:kg}).
Finally, we prompt GPT-4 with the knowledge triples and a specific scenario to generate dialogue.
For example, if \textit{Reno} has low health and is using \texttt{EM shot} (right, Figure~\ref{fig:steps}).
We further discuss the prompt process specifically for \texttt{FFVIIR} and Pokémon in Section~\ref{sec:prompt_ffvii} and Section~\ref{sec:prompt_pokemon}, respectively.

\subsection{Knowledge Graph Acquisitions}
\label{sec:kg}

\textbf{Crawling:} We begin by crawling data from fandom websites of Final Fantasy VII and Pokémon as mentioned above. In Final Fantasy VII Remake, we gather detailed information on characters, including their personalities, abilities, relationships, and the events they influence or participate in. Additionally, we compile comprehensive data on boss battles, covering skills, weaknesses, and various states, such as low health or being engulfed in flames. 
In Pokémon, we compiled comprehensive data on the personality traits, appearances, and Pokémon ownership of all main NPCs in the game. Additionally, we utilized the existing Pokémon knowledge graph, which encompasses detailed information about individual Pokémon, their attributes, evolutions, and relationships. This data forms the foundation for creating AI dialogue that can utilize knowledge-based to enhance player immersion.



\textbf{Knowledge Graph Triples: } From the dataset and ontology developed in previous steps, we have constructed small knowledge graphs tailored to FFVIIR and Pokémon game. These knowledge graphs are designed to store and organize the information about non-player characters (NPCs) and their associated Pokémon, facilitating efficient information retrieval and enhancing the interaction experience in gameplay.

\begin{figure}[h]
    \begin{quote}
        ``\textit{Sabrina is a female character, a slim young woman of medium height. She wears a small red and dress black in the middle at the waist. Pokémon: Mr. Mime ...}''
    \end{quote}
    \hspace*{3.5cm}$\downarrow$
    \vspace{3mm}\\
    \small{ \texttt{
    \hspace*{8mm}(Sabrina, has\_gender, female)\\
    \hspace*{8mm}(Sabrina, has\_outfit, a small red and \\
    \hspace*{8mm} dress black in the middle at the waist)\\
    \hspace*{8mm}(Sabrina, has\_height, slim young woman)\\
    \hspace*{8mm}(Sabrina, has\_pokemon, Mr. Mime)}}
    \caption{This figure demonstrates the conversion of a descriptive text about Sabrina, a character from the game, into RDF triples. Key attributes such as gender, outfit, height, and associated Pokémon are extracted and formatted as triples, showcasing the methodology for semantic representation of character information.}
    \label{fig:sabrina_triple}
\end{figure}

Figure~\ref{fig:sabrina_triple} shows an example of Pokémon KG where the KG contains a set of triples, structured as subject-predicate-object, that describe the relationships and attributes of NPCs and Pokémon within the game's universe. By integrating these triples into a knowledge graph, we can utilize these triples in LLMs' to generate context-aware dialogues based on the NPCs' distinct personalities, appearances, and relationships with their Pokémon.


\subsection{Prompting Methods: Final Fantasy VII Remake}
\label{sec:prompt_ffvii}
\begin{figure}[t]
    \centering
    \includegraphics[width=0.4\textwidth]{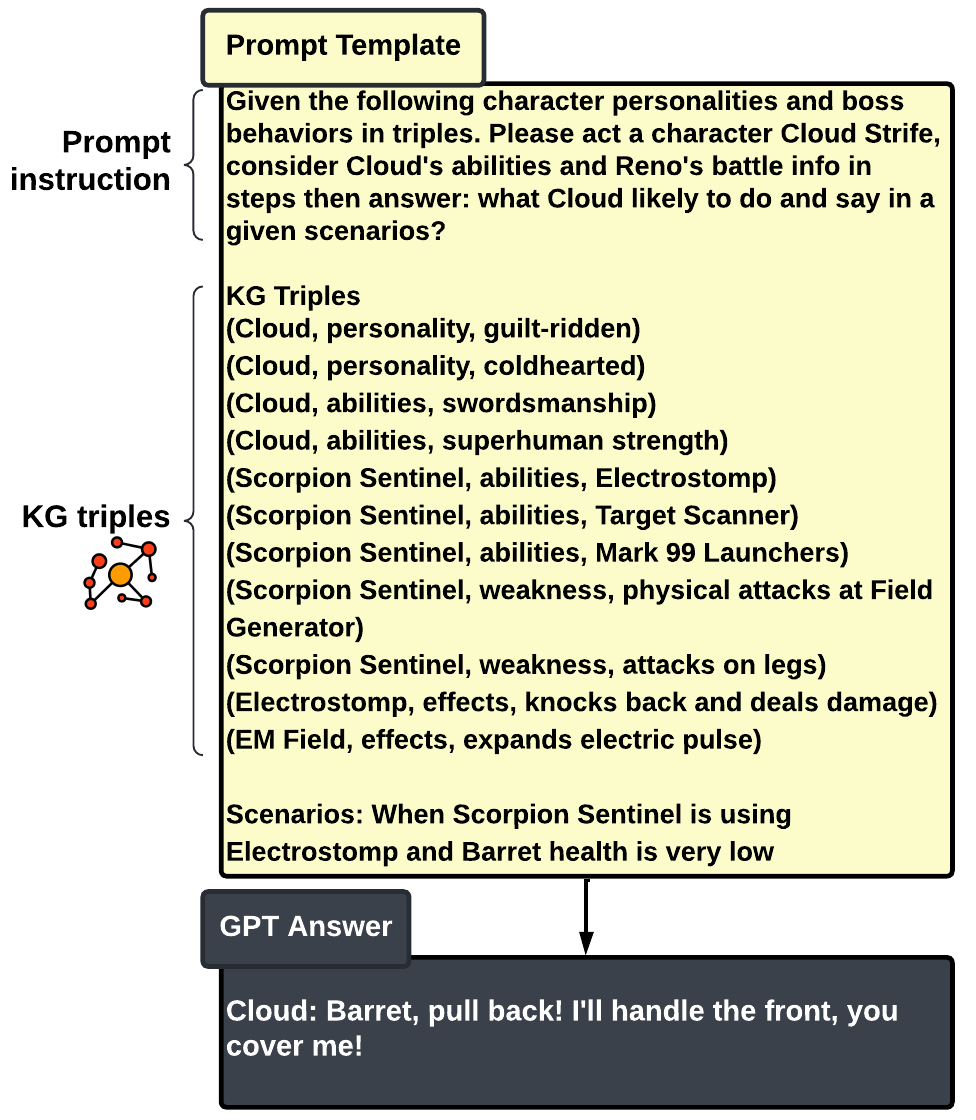}
    \caption{An illustrative example of the complete prompt for dialogue generation in \texttt{FFVIIR} battle. Cloud personalities and Scorpion Sentinel (boss) battle information in knowledge graph (KG) triples are provided and used to generate context-aware dialogue. The process involves considering a battle situation when Cloud is facing Scorpion Sentinel.}
    \label{fig:prompt-ffvii}
\end{figure}

Though \texttt{FFVIIR} delivers cutscenes with story-driven dialogue and visually stunning graphics, its battle gameplay can become lengthy during some boss fights, potentially tedious players.
To make the in-battle dialogue more dynamic, we prompt GPT-4 using knowledge triples, as discussed in Section~\ref{sec:kg}, to generate dialogue that responds to specific scenarios.
In each prompt, we provide: (1) an instruction, (2) character triples, (3) boss triples, and (4) a specific situation in a battle (Figure ). 
We ensure that the instruction guides GPT-4 to use the given knowledge about characters and bosses.
For example (Figure~\ref{fig:prompt-ffvii}), Cloud can be cold but tactical. So, when Barret is low on health and the \texttt{Scorpion Sentinel} is about to use \textit{Electrostomp} (a wide-range electric attack around its body), Cloud would tell Barret to pull back and use \textit{cure}.

\begin{figure}[t]
    \centering
    \includegraphics[width=0.4\textwidth]{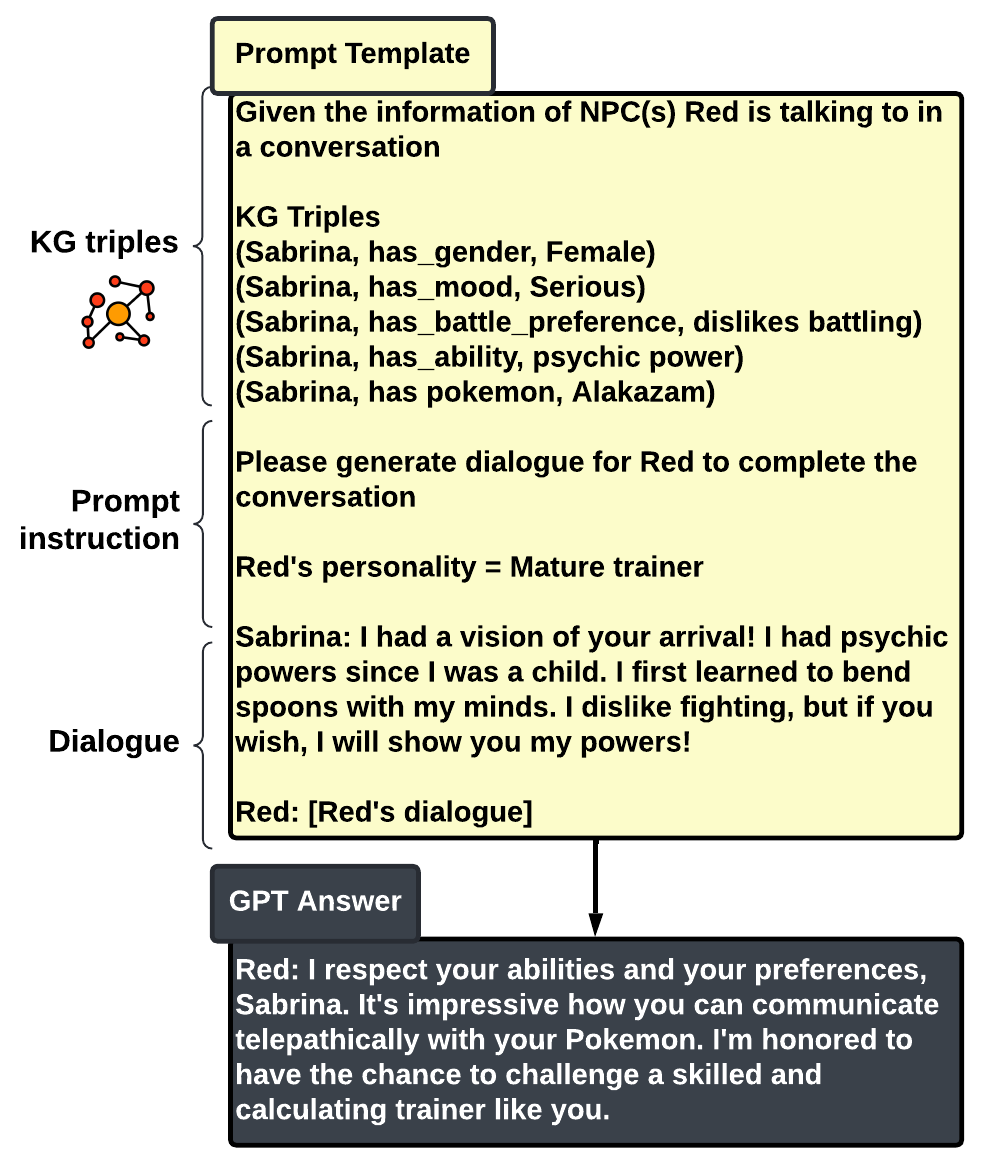}
    \caption{An illustrative example of the complete prompt used for dialogue generation. The character-specific information from a knowledge graph (KG) is converted into KG triples and used to generate context-aware dialogue. The process involves concatenating actual dialogue from Sabrina, the character with whom Red interacts in the game.}
    \label{fig:prompt-pokemon}
\end{figure}

\subsection{Prompting Methods: Generating Red's responses to NPCs}
\label{sec:prompt_pokemon}

In the original Pokémon games, the protagonist—known as Red or Ash—does not engage in verbal interactions with NPCs. Instead, the gameplay is structured such that players only see text responses from the NPCs during these interactions. To enhance this dynamic, we have devised a prompt template that merges our knowledge graph data with authentic Pokémon scripts. This integration facilitates querying the knowledge graph for concepts related to the main character’s interactions, a strategy aimed at optimizing in-game exchanges from a game development standpoint, where comprehensive data is utilized to craft the most engaging interactions with NPCs. Figure \ref{fig:prompt-pokemon} illustrates a prompt example, depicting a hypothetical conversation between Red and the Gym Leader Sabrina.

%% file: sections/40-results.tex
\section{Preliminary results}
On the surface, GPT-4 looks promising for text generation. 
However, can it generate dialogue in a manner similar to humans? 
We aim to answer this question with both qualitative and quantitative results.
Currently, our results are limited to qualitative assessments (Section~\ref{sec:result_ffvii} and Section~\ref{sec:result_pokemon}).
We highlight words in responses as follows:
\begin{itemize}
    \item in \colorbox{cyan}{blue} if that reflects GPT-4 majorly using given knowledge triples
    \item in \colorbox{lime}{green} if that reflects GPT-4 majorly confronting a situation 
\end{itemize}
We plan to conduct human evaluations to obtain quantitative data, which we will discuss in Section~\ref{sec:result_quan}.

\subsection{Cloud positively fights \texttt{FFVIIR} bosses.}
\label{sec:result_ffvii}
In \texttt{FFVIIR}, there are 24 bosses that players must fight.
Due to time and resource constraints, we select a subset of these bosses to sample Cloud's responses in specific situations.
We choose the first boss of \texttt{FFVIIR} (\texttt{Scorpion Sentinel}), \texttt{Reno}, and the last boss, \texttt{Sephiroth}, to show how GPT-4 acts as Cloud responses during the battles. 
We sample GPT-4 to generate five responses and pick the best one to show in Table~\ref{tab:ffvii_boss_scenarios} with variations of battle situations in each selected boss.
In total, we have 3 bosses, 13 situations, and 65 responses from GPT-4. 

\textbf{Cloud is overly positive in fights}.
While this might contradict players' experiences, it demonstrates how GPT-4 (or any large language models) tend to generalize positive traits, even when provided with examples portraying Cloud as cold-hearted and arrogant.
With the next step, we hope to supervise and fine-tune GPT-4 to align more closely with Cloud's true personality.

\textbf{GPT-4 responds based on situations rather than knowledge.}
Since we instruct GPT-4 to focus on Cloud's abilities and the bosses' weaknesses, we expect it to respond with more factual knowledge. However, we notice that while some examples do leverage knowledge of bosses' skills and weaknesses (highlighted in \colorbox{cyan}{blue} in Table~\ref{tab:ffvii_boss_scenarios}), the majority of the responses are based on situational context (highlighted in \colorbox{lime}{green} in Table~\ref{tab:ffvii_boss_scenarios}), including some that were omitted.

\begin{table*}[ht]
\centering
\begin{tabular}{|c|p{6cm}|p{6cm}|}
 \hline
  \textbf{Boss} & \textbf{Scenarios} & \textbf{Cloud's responses (Best of 5)} \\
 \hline
 \multirow{7}{*}{{Scorpion Sentinel}}  &\footnotesize{[When Scorpion Sentinel is using Electrostomp and Barret health is very low]} &\footnotesize{\textit{"\colorbox{lime}{Barret, pull back!} I'll handle the front, you cover me!"}} \\
&\footnotesize{[When Scorpion Sentinel is using Electrostomp and Barret health is normal]} &\footnotesize{\textit{"\colorbox{lime}{Barret, keep its attention.} I'm going in for the kill."}}\\
&\footnotesize{[When Scorpion Sentinel engages its Tail Laser]} &\footnotesize{\textit{"You're not \colorbox{lime}{catching} me with that one!"}}\\
&\footnotesize{[When Scorpion Sentinel first activates its Auto-Repair]}&\footnotesize{\textit{"It's \colorbox{lime}{healing up!} Focus attacks on \colorbox{cyan}{the core}—now!"}}\\
&\footnotesize{[Upon defeating the Scorpion Sentinel]}&\footnotesize{\textit{"Looks like your \colorbox{cyan}{auto-repair} took the day off. Bad timing."}}\\
 \hline
 \multirow{4}{*}{{Reno}}  &\footnotesize{[After Reno prepares EM Flail]}&\footnotesize{\textit{"Not getting \colorbox{lime}{caught} in that mess. Let's see how you handle this!"}}\\
&\footnotesize{[When Reno uses EM Shot]}&\footnotesize{\textit{"You’re just making this easier for me."}}\\
&\footnotesize{[Before Reno uses EM Mine Toss]}&\footnotesize{\textit{"You really think \colorbox{lime}{those toys} will stop me?"}}\\
&\footnotesize{[After defeating Reno]}&\footnotesize{\textit{"Reno, relying too much on those \colorbox{cyan}{EM Mines} was a bad call. \colorbox{cyan}{They're too predictable.}"}}\\
 \hline
 \multirow{4}{*}{{Sephiroth}}  &\footnotesize{[When the battle against Sephiroth begins, he starts casting Firaga]}&\footnotesize{\textit{"Firaga, huh? Guess it's time to \colorbox{cyan}{chill} things down."}}\\
&\footnotesize{[After Sephiroth falls to 80\% health and using Thunderstorm]}&\footnotesize{\textit{"\colorbox{lime}{This storm}’s just a breeze!"}}\\
&\footnotesize{[After Sephiroth falls to 30\% health and using Boundless Void]}&\footnotesize{\textit{"This is for \colorbox{cyan}{everything you've done!}"}}\\
&\footnotesize{[After defeating Sephiroth though he just walk away with no scratches]}&\footnotesize{\textit{"No matter how many times we do this, \colorbox{cyan}{you keep coming back...} But so will I. I won't stop until it's finished, for good."}}\\
 \hline
\end{tabular}
\caption{Examples of GPT-4 generating dialogue for Cloud, who is in a battle fighting boss: Scorpion Sentinel, Reno, and Sephiroth. In each boss battle, Cloud can respond to multiple scenarios. We sample 5 responses for each scenario and select the best one to illustrate.}
\label{tab:ffvii_boss_scenarios}
\end{table*}

\subsection{Pokemon's Red responses to NPCs.}
\label{sec:result_pokemon}

We conducted the experiment as outlined in Section \ref{sec:prompt_pokemon}, utilizing a prompting method tailored for the Pokémon game context. To explore the impact of personality on dialogue dynamics, we endowed the main character Red with five distinct personas: (1) mature Pokémon trainer, (2) amateur Pokémon trainer, (3) talkative, (4) timid, and (5) confident. We carefully selected prominent NPCs for this experiment, including all eight gym leaders, the elite four, Professor Oak, and Rival Blue, to ensure a comprehensive evaluation across varied and significant game interactions.

\begin{figure}[t]
    \centering
    \includegraphics[width=0.49\textwidth]{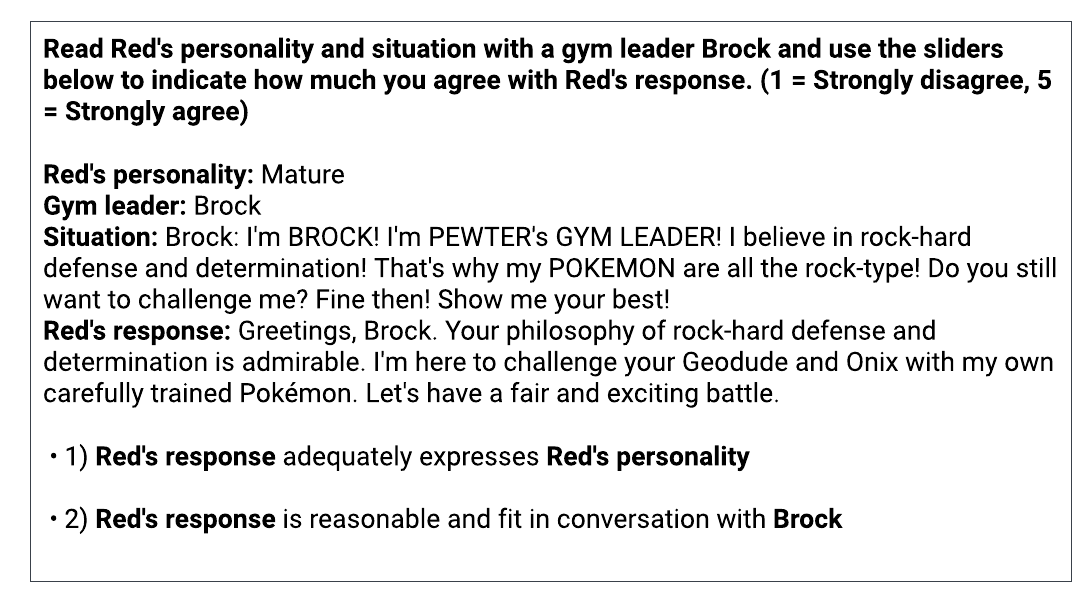}
    \caption{An illustration example of a human evaluation task created from a conversation between Red and Brock.}
    \label{fig:eval_1}
\end{figure}

\begin{figure}[t]
    \centering
    \includegraphics[width=0.49\textwidth]{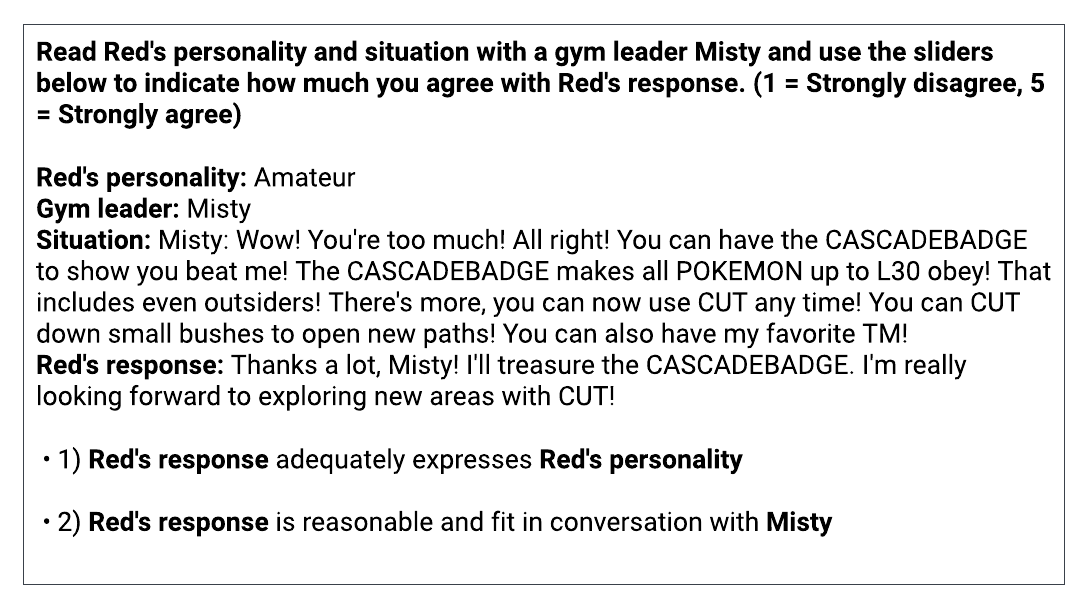}
    \caption{An illustration example of a human evaluation task created from a conversation between Red and Misty.}
    \label{fig:eval_2}
\end{figure}

In total, we generated 120 dialogues for Red, utilizing GPT-4 to synthesize responses based on the specified personalities. Though our resources are limited, we conduct a human evaluation of these dialogues by instruction and providing two main tasks for the evaluator:  
\begin{itemize}
    \item \textbf{Instruction:} Read Red's personality and situation with a gym leader and use the sliders below to indicate how much you agree with Red's response (1 = Strongly disagree, 5 = Strongly agree).
    \item \textbf{Given:} Red's personality, a gym leader's name, situation, and Red's response. 
    \item \textbf{Task 1:} Red's response adequately expresses Red's personality
    \item \textbf{Task 2:} Red's response is reasonable and fits in conversation with a gym leader
\end{itemize}
One example from the human evaluation is shown in \ref{fig:eval_1} and \ref{fig:eval_2}, in which a total of 120 responses are evaluated.

\begin{figure}[t]
    \centering
    \includegraphics[width=0.47\textwidth]{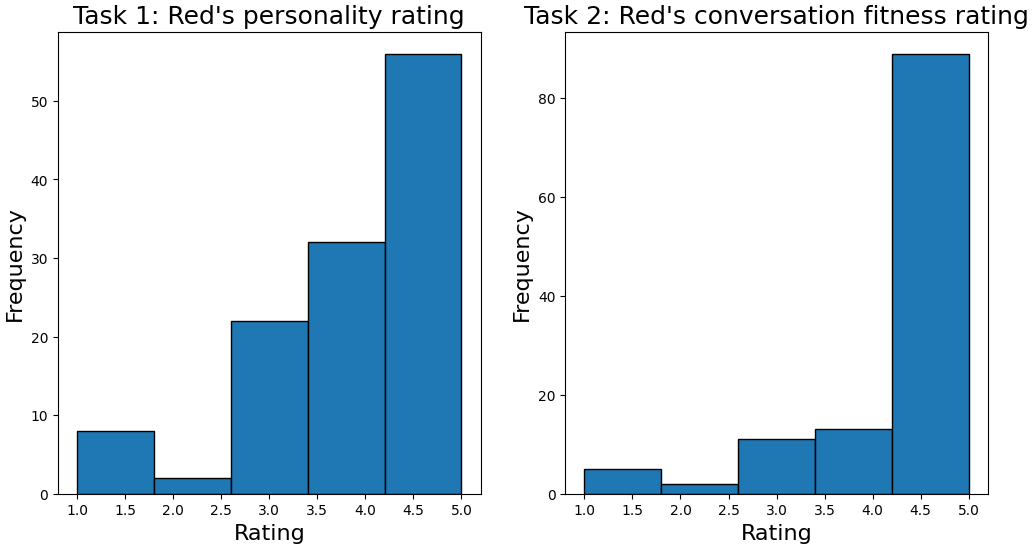}
    \caption{Histograms illustrate the distribution of ratings for how well Red's responses reflect his personality (left) and Red's conversation fitness (right).}
    \label{fig:human-eval}
\end{figure}

\begin{table*}[ht]
\centering
\renewcommand{\arraystretch}{1.2}  
\setlength{\tabcolsep}{5pt}  
\begin{tabular}{|c|p{12cm}|}
\hline
\multicolumn{2}{|c|}{\parbox{12cm}{\vspace{1.5mm}"Brock: I'm BROCK! I'm PEWTER's GYM LEADER! I believe in rock-hard defense and determination! That's why my POKEMON are all the rock-type! Do you still want to challenge me? Fine then! Show me your best!"}} \\ \hline
\textbf{Personality} & \textbf{Red's Response} \\ \hline
Mature Pokémon Trainer & \textit{"Greetings, Brock. Your philosophy of rock-hard defense and determination is admirable. I'm here to challenge your \colorbox{cyan}{Geodude and Onix} with my own carefully trained Pokémon. Let's have a fair and exciting battle."} \\ \hline
Amateur Pokémon Trainer & \textit{"Wow, Brock! I’ve heard a lot about you and your rock-solid Pokémon, like \colorbox{cyan}{Geodude and Onix}. I’m ready to give it my best shot and learn from this battle. Let’s do this!"} \\ \hline
Talkative & \textit{"Hey, Brock! Wow, you look really cool with that outfit and your rock-hard Pokémon team! I’ve heard so much about your \colorbox{cyan}{Geodude and Onix}. I can’t wait to see how my Pokémon stack up against your rock-type defense. Let's have an awesome battle!"} \\ \hline
Timid & \textit{"Um, h-hi Brock. Y-yes, I’d like to challenge you. I know your \colorbox{cyan}{Geodude and Onix} are really strong, but I’ll do my best. Let’s, uh, begin the battle."} \\ \hline
Confident & \textit{"Brock, I've been looking forward to this challenge. Your \colorbox{cyan}{Geodude and Onix} are legendary for their defense, but I'm confident my team is up to the task. Let's have a great battle!"} \\ \hline
\end{tabular}
\caption{Examples of GPT-4 generated dialogue for a conversation between Red and Brock, the first gym leader of Pokémon Red version. The examples show five different responses from five personality of the main character Red.}
\label{tab:personality_result}
\end{table*}

\textbf{GPT-4 is able to generate personality-based dialogue: }The results of our additional experiment revealed significant differences in dialogue generation based on the specified personalities. The experiment's outcomes demonstrate GPT-4's remarkable ability to generate distinctly varied dialogue that aligns with predefined personality profiles. The quality of GPT-4 Red's responses for \textbf{Task 1}, though not perfect, are high-quality and show keywords corresponding to a given Red's personality. Figure \ref{fig:human-eval} (Left) shows that more than 50 of 120 responses are rated with scores of 4.5-5.0 while around 30 are rated with scores of 3.5-4.5, 20 are acceptable (2.5-3.5), and around ten are low-quality (under 2.5).

\begin{figure}[t]
    \centering
    \includegraphics[width=0.30\textwidth]{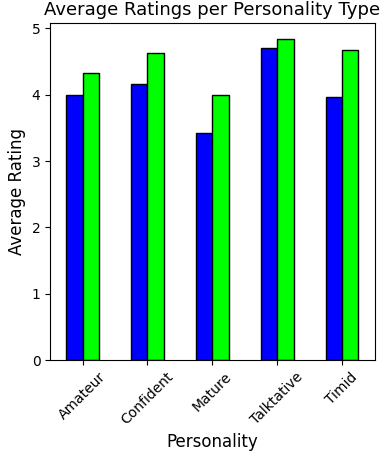}
    \caption{A bar chart displays the average ratings based on each personality type. The blue bars represent the average personality ratings, while the lime bars represent the average conversation fitness ratings.}
    \label{fig:human-eval-person}
\end{figure}

As shown Table \ref{tab:personality_result}, GPT-4 not only adapts its responses based on the personality trait assigned to the main character Red but also effectively incorporates specific information from the knowledge graph. For instance, the model's ability to mention specific Pokémon, such as Geodude and Onix, in its responses. The same patterns can be observed when generating dialogue for Red with other NPCs in the Pokémon game where  (\textbf{Task 2}, Figure \ref{fig:human-eval}).

\textbf{GPT-4 is good at personating a talkative Red}. From all personalities given for Red, GPT-4 retrieves the highest score for \textbf{Task 1} and \textbf{Task 2} when Red is talkative (Figure \ref{fig:human-eval-person}). Responses, when Red is a mature trainer, are rated the lowest for both tasks. From this analysis, straightforward personalities such as talkative, timid, and confident are trivial for GPT-4 to derive, while those with unclear personalities, like being mature, might be challenging for GPT-4 to clearly express in words.

\subsection{Future work}
\label{sec:result_quan}
Initially, we want to see how well GPT-4 can act as a game character and provide responses that are unexpected yet fitting.
We aim to expand to other video game series that are less popular than the two we selected. In addition, we want to conduct more human evaluations for those games, including Final Fantasy VII Remake.
For Final Fantasy VII Remake (\texttt{FFVIIR}) that has personality specifically, we will present multiple character information with situations accompanied by GPT-4's responses. 
Participants will then have $K$ options to choose the character that they believe best fits the given responses. 
A high \textit{F}-score would indicate that GPT-4 maintains consistent character traits.

Additionally, we want to improve the evaluation to prove how well GPT-4 uses knowledge in its responses. 
In this task, participants will highlight words in the responses: (1) in blue if the response reflects GPT-4 predominantly using given knowledge, and (2) in green if it reflects GPT-4 primarily responding to the situation (similar to our approach in  in Table~\ref{tab:ffvii_boss_scenarios} and Table~\ref{tab:personality_result}).

These additional evaluations show great promise and are valid for implementation and broader application in dialogue generation tasks and multiple game scenarios.

%% file: sections/50-related.tex
\section{Related works}

\textbf{Dialogue and Text Generation in Games:} Text-based games have proven to be excellent test beds for the development of agents operating within linguistic frameworks. As these games involve interactions such as input observations and action commands that are entirely text-driven, they demand a thorough understanding of language to facilitate successful gameplay. Prior studies have explored various text generation techniques in such gaming environments. For instance, \cite{fulda2017can} utilized Word2vec embeddings to introduce inference capabilities, enabling the generated text to learn associations between suitable verbs and objects. Further research \cite{urbanek2019learning} extended this approach by training BERT \cite{devlin2018bert} to produce contextually relevant dialogue in action and fantasy game settings.

\textbf{Knowledge Graph enhancing Text Generation: }Knowledge graphs have become an reliable resource for knowledge-grounded conversation across various research domains \cite{tuan2019dykgchat,kang2022knowledge}. They are instrumental in enhancing dialogue generation in fields such as medicine \cite{varshney2023knowledge}, question-answering systems \cite{indurthi2017generating}, and gaming \cite{ashby2023personalized}. By integrating KGs, these systems are able to leverage structured knowledge to improve the relevance and accuracy of generated text, significantly boosting performance.

\textbf{Generation in Text-based Games using LLMs:} Large language models (LLMs) have been increasingly employed to enhance and innovate text-based games. \cite{ammanabrolu2019toward} leveraged Markov chains alongside neural language models to dynamically generate quests in AI Dungeon 2 \cite{walton2020ai}, expanding gameplay complexity and narrative depth. Building on this, a recent approach outlined in \cite{NEURIPS2023_a9bbeb28} introduced MarioGPT, an LLM-based text-to-level model that enables the creation of diverse game levels tailored to different play styles, further demonstrating the adaptability of LLMs in gaming. Additionally, in the realm of narrative generation, \cite{kumaran2023scenecraft} proposed a novel application of LLMs for generating narrative scenes in digital games, enhancing the storytelling aspect significantly. These advancements highlight the transformative potential of LLMs in redefining gaming experiences by providing rich, adaptive, and personalized content

%% file: sections/60-conclusion.tex
\section{Conclusion}
In this study, we have explored the capacity of GPT-4 to accurately emulate game characters and provide responses that align with their personalities and situational contexts. By focusing on main characters of two popular game Final Fantasy VII and Pokémon, we investigated how well GPT-4 can maintain character consistency and utilize in-game knowledge when generating dialogue.

Our qualitative suggests  that GPT-4 can maintain a high degree of character consistency. Additionally, the task of highlighting knowledge-based and situational responses has provided further insights into the model's ability to integrate specific game details into its responses.

Future steps involve refining the model through supervised fine-tuning to better align GPT-4's responses with both of the main characters personality traits. Furthermore, we aim to extend our evaluation methods to human evaluation and apply it to other characters and dialogue generation tasks within various game contexts. These efforts will help enhance the interactive experience for players, ensuring that AI-generated dialogues remain immersive and true to character.

Overall, this study demonstrates the potential of GPT-4 in enhancing character-driven interactions in gaming, paving the way for more sophisticated and engaging LLMs-driven dialogue systems.

%% file: main.bbl
\begin{thebibliography}{15}
\expandafter\ifx\csname natexlab\endcsname\relax\def\natexlab#1{#1}\fi

\bibitem[{Ammanabrolu et~al.(2019)Ammanabrolu, Broniec, Mueller, Paul, and Riedl}]{ammanabrolu2019toward}
Prithviraj Ammanabrolu, William Broniec, Alex Mueller, Jeremy Paul, and Mark~O Riedl. 2019.
\newblock Toward automated quest generation in text-adventure games.
\newblock \emph{arXiv preprint arXiv:1909.06283}.

\bibitem[{Ashby et~al.(2023)Ashby, Webb, Knapp, Searle, and Fulda}]{ashby2023personalized}
Trevor Ashby, Braden~K Webb, Gregory Knapp, Jackson Searle, and Nancy Fulda. 2023.
\newblock Personalized quest and dialogue generation in role-playing games: A knowledge graph-and language model-based approach.
\newblock In \emph{Proceedings of the 2023 CHI Conference on Human Factors in Computing Systems}, pages 1--20.

\bibitem[{Biswas et~al.(2021)Biswas, Sofronova, Alam, and Sack}]{biswas2021contextual}
Russa Biswas, Radina Sofronova, Mehwish Alam, and Harald Sack. 2021.
\newblock Contextual language models for knowledge graph completion.
\newblock In \emph{MLSMKG@ PKDD/ECML}, page~13.

\bibitem[{Devlin et~al.(2018)Devlin, Chang, Lee, and Toutanova}]{devlin2018bert}
Jacob Devlin, Ming-Wei Chang, Kenton Lee, and Kristina Toutanova. 2018.
\newblock Bert: Pre-training of deep bidirectional transformers for language understanding.
\newblock \emph{arXiv preprint arXiv:1810.04805}.

\bibitem[{Fulda et~al.(2017)Fulda, Ricks, Murdoch, and Wingate}]{fulda2017can}
Nancy Fulda, Daniel Ricks, Ben Murdoch, and David Wingate. 2017.
\newblock What can you do with a rock? affordance extraction via word embeddings.
\newblock \emph{arXiv preprint arXiv:1703.03429}.

\bibitem[{Indurthi et~al.(2017)Indurthi, Raghu, Khapra, and Joshi}]{indurthi2017generating}
Sathish~Reddy Indurthi, Dinesh Raghu, Mitesh~M Khapra, and Sachindra Joshi. 2017.
\newblock Generating natural language question-answer pairs from a knowledge graph using a rnn based question generation model.
\newblock In \emph{Proceedings of the 15th Conference of the European Chapter of the Association for Computational Linguistics: Volume 1, Long Papers}, pages 376--385.

\bibitem[{Kang et~al.(2022)Kang, Kwak, Baek, and Hwang}]{kang2022knowledge}
Minki Kang, Jin~Myung Kwak, Jinheon Baek, and Sung~Ju Hwang. 2022.
\newblock Knowledge-consistent dialogue generation with knowledge graphs.
\newblock In \emph{ICML 2022 Workshop on Knowledge Retrieval and Language Models}.

\bibitem[{Kumaran et~al.(2023)Kumaran, Rowe, Mott, and Lester}]{kumaran2023scenecraft}
Vikram Kumaran, Jonathan Rowe, Bradford Mott, and James Lester. 2023.
\newblock Scenecraft: automating interactive narrative scene generation in digital games with large language models.
\newblock In \emph{Proceedings of the AAAI Conference on Artificial Intelligence and Interactive Digital Entertainment}, volume~19, pages 86--96.

\bibitem[{OpenAI(2024)}]{openai2024gpt4}
OpenAI. 2024.
\newblock \href {http://arxiv.org/abs/2303.08774} {Gpt-4 technical report}.

\bibitem[{Pan et~al.(2024)Pan, Luo, Wang, Chen, Wang, and Wu}]{pan2024unifying}
Shirui Pan, Linhao Luo, Yufei Wang, Chen Chen, Jiapu Wang, and Xindong Wu. 2024.
\newblock Unifying large language models and knowledge graphs: A roadmap.
\newblock \emph{IEEE Transactions on Knowledge and Data Engineering}.

\bibitem[{Sudhakaran et~al.(2023)Sudhakaran, Gonz\'{a}lez-Duque, Freiberger, Glanois, Najarro, and Risi}]{NEURIPS2023_a9bbeb28}
Shyam Sudhakaran, Miguel Gonz\'{a}lez-Duque, Matthias Freiberger, Claire Glanois, Elias Najarro, and Sebastian Risi. 2023.
\newblock \href {https://proceedings.neurips.cc/paper_files/paper/2023/file/a9bbeb2858dfbdbd4c19814e5d80ec60-Paper-Conference.pdf} {Mariogpt: Open-ended text2level generation through large language models}.
\newblock In \emph{Advances in Neural Information Processing Systems}, volume~36, pages 54213--54227. Curran Associates, Inc.

\bibitem[{Tuan et~al.(2019)Tuan, Chen, and Lee}]{tuan2019dykgchat}
Yi-Lin Tuan, Yun-Nung Chen, and Hung-yi Lee. 2019.
\newblock Dykgchat: Benchmarking dialogue generation grounding on dynamic knowledge graphs.
\newblock \emph{arXiv preprint arXiv:1910.00610}.

\bibitem[{Urbanek et~al.(2019)Urbanek, Fan, Karamcheti, Jain, Humeau, Dinan, Rockt{\"a}schel, Kiela, Szlam, and Weston}]{urbanek2019learning}
Jack Urbanek, Angela Fan, Siddharth Karamcheti, Saachi Jain, Samuel Humeau, Emily Dinan, Tim Rockt{\"a}schel, Douwe Kiela, Arthur Szlam, and Jason Weston. 2019.
\newblock Learning to speak and act in a fantasy text adventure game.
\newblock \emph{arXiv preprint arXiv:1903.03094}.

\bibitem[{Varshney et~al.(2023)Varshney, Zafar, Behera, and Ekbal}]{varshney2023knowledge}
Deeksha Varshney, Aizan Zafar, Niranshu~Kumar Behera, and Asif Ekbal. 2023.
\newblock Knowledge graph assisted end-to-end medical dialog generation.
\newblock \emph{Artificial Intelligence in Medicine}, 139:102535.

\bibitem[{Walton(2020)}]{walton2020ai}
Nick Walton. 2020.
\newblock Ai dungeon 2: creating infinitely generated text adventures with deep learning language models.

\end{thebibliography}
